\documentclass[conference]{IEEEtran}
\IEEEoverridecommandlockouts
\usepackage{cite}
\usepackage{amsmath,amssymb,amsfonts}
\usepackage{algorithmic}
\usepackage{multirow}
\usepackage[subrefformat=parens]{subcaption}
\usepackage{comment}
\usepackage[dvipdfmx]{graphicx}
\usepackage{textcomp}
\usepackage{url}
\usepackage{xcolor}
\def\BibTeX{{\rm B\kern-.05em{\sc i\kern-.025em b}\kern-.08em
    T\kern-.1667em\lower.7ex\hbox{E}\kern-.125emX}}
\renewcommand\IEEEkeywordsname{Keywords}

\makeatletter
\def\ps@IEEEtitlepagestyle{%
  \def\@oddfoot{\mycopyrightnotice}%
  \def\@evenfoot{}%
}
\def\mycopyrightnotice{%
  \begin{minipage}{\textwidth}
  \centering \scriptsize
  Copyright~\copyright~2022 IEEE. Personal use of this material is permitted.  Permission from IEEE must be obtained for all other uses, in any current or future media, including reprinting/republishing this material for advertising or promotional purposes, creating new collective works, for resale or redistribution to servers or lists, or reuse of any copyrighted component of this work in other works.
  \end{minipage}
}
\makeatother
\begin{document}

\title{Conservative Likelihood Ratio Estimator for Infrequent Data Slightly above a Frequency Threshold}

\author{
\IEEEauthorblockN{Masato Kikuchi\IEEEauthorrefmark{1}, Yuhi Kusakabe\IEEEauthorrefmark{1}, Tadachika Ozono}
\IEEEauthorblockA{\textit{Department of Computer Science, Graduate School of Engineering} \\
\textit{Nagoya Institute of Technology}\\
Nagoya, Aichi, Japan \\
kikuchi@nitech.ac.jp, kusayu@ozlab.org, ozono@nitech.ac.jp \\
\IEEEauthorrefmark{1} The first and second authors contributed equally to this study.}
}

\maketitle

\begin{abstract}
A naive likelihood ratio (LR) estimation using the observed frequencies of events can overestimate LRs for infrequent data.
One approach to avoid this problem is to use a frequency threshold and set the estimates to zero for frequencies below the threshold.
This approach eliminates the computation of some estimates, thereby making practical tasks using LRs more efficient.
However, it still overestimates LRs for low frequencies near the threshold.
This study proposes a conservative estimator for low frequencies, slightly above the threshold.
Our experiment used LRs to predict the occurrence contexts of named entities from a corpus.
The experimental results demonstrate that our estimator improves the prediction accuracy while maintaining efficiency in the context prediction task.
\end{abstract}

\begin{IEEEkeywords}
\textit{\textbf{likelihood ratio, conservative estimation, frequency threshold, infrequent data, efficiency}}
\end{IEEEkeywords}

\section{Introduction}
\label{sec:intro}

Likelihood ratios (LRs) are well-used statistics in statistical tests~\cite{Glover:04} and binary classification~\cite{Nakanishi:15}, and the estimation of LRs significantly impacts the effectiveness of applications that use them.
In natural language processing, LRs are often estimated based on the observed frequencies of discrete elements, such as letters or words in a corpus~\cite{Dunning:93,Manning:99}.
Suppose that we estimate the LR as follows:
\begin{align*}
r(x) = \frac{p_{\rm nu}(x)}{p_{\rm de}(x)},
\end{align*}
where $x$ denotes a discrete element.
The indices ``de'' and ``nu'' represent the denominator and numerator of $r(x)$, respectively.
A naive estimation approach estimates the probability distributions $p_{*}(x)$ as the relative frequencies $\widehat{p}_{*}(x)$, $* \in \{{\rm de, nu}\}$ and takes their ratio.
The estimator $r_{\rm MLE}(x)$ is defined as follows:
\begin{align}
\label{eq:MLE}
\widehat{p}_{*}(x) = \frac{f_{*}(x)}{n_{*}}, \quad
r_{\rm MLE}(x) = \frac{\widehat{p}_{\rm nu}(x)}{\widehat{p}_{\rm de}(x)},
\end{align}
where $f_{*}(x)$ denotes the observed frequency of $x$ sampled from the probability distribution following the density $p_{*}(x)$, and $n_{*}$ is $\sum_x f_{*}(x)$.
Because language resources contain a limited number of elements and their frequency distribution follows a power law, there are several low-frequency elements.
In this case, the naive estimation approach has the problem of overestimating LRs of the low-frequency elements.
One approach to avoid this problem is to use a frequency threshold and set the estimates to zero for frequencies below the threshold.
The estimator $r_{\rm th}(x)$ is defined as follows:
\begin{align}
\label{eq:th}
r_{\rm th}(x) = \begin{cases}
	r_{\rm MLE}(x) & {\rm if\ } f_{\rm nu}(x) >  \theta_{\rm th}, \\
	0 & {\rm otherwise},
	\end{cases}
\end{align}
where $\theta_{\rm th}\ (\ge 0)$ denotes the frequency threshold.
For convenience, this estimator prevents the overestimation of LRs for frequencies below $\theta_{\rm th}$.
Furthermore, it can improve the efficiency of practical tasks that use LRs because it eliminates LR estimation of infrequent elements.
However, because this estimator uses $r_{\rm MLE}(x)$ in its definition, it still overestimates LRs for low frequencies, slightly above the threshold.

This study proposes a conservative estimator for low frequencies near the threshold.
Our estimator conservatively (low) estimates the LRs of low-frequency elements slightly above a threshold, and uniformly sets the LRs of elements below the threshold to zero.

\begin{figure}[tb]
  \centering
  \includegraphics[keepaspectratio, scale=0.62]{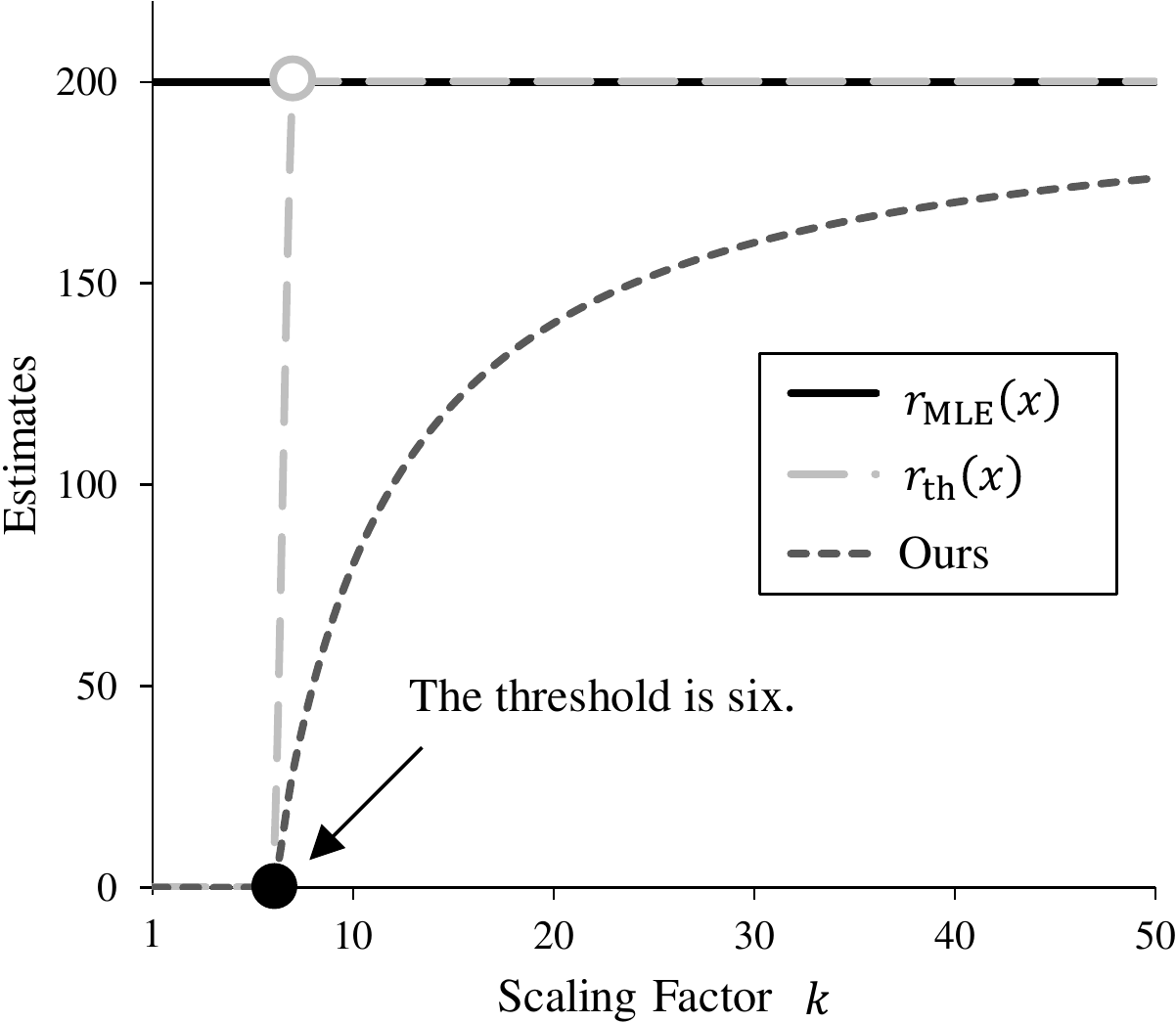}
  \caption{Behaviors of each estimator.
  We assume that $r_{\rm MLE}(x)$ is constant ($200$) regardless of $k$.
  ``Ours'' indicates our proposed estimator.}
  \label{fig:behavioral_analysis}
\end{figure}
For further understanding, we used LR estimation examples for each estimator as we varied the observed frequencies.
We define the frequencies of $x$ as follows:
\begin{align*}
&n_{\rm de} = 10^7,\ f_{\rm de}(x) = 5 \times k, \\
&n_{\rm nu} = 10^4,\ f_{\rm nu}(x) = k,
\end{align*}
where $k$ is a scaling factor that takes any natural number, and we adjust it to vary these frequencies.
We estimate $r(x) $ based on the frequencies.
When $k$ is large, $f_{\rm de}(x)$ and $f_{\rm nu}(x)$ occur frequently.
By contrast, when $k$ is small, they occur infrequently.
Fig.~\ref{fig:behavioral_analysis} shows the behaviors of each estimator when varying $k$ from 1 to 50.
The horizontal axis represents the scaling factor $k$, and the vertical axis represents the estimates corresponding to $k$.
We set a common threshold of six for $r_{\rm th}(x)$ and our estimator.
Note that $r_{\rm MLE}(x)$ is $200$ regardless of $k$ as follows:
\begin{align*}
r_{\rm MLE}(x) = \frac{\widehat{p}_{\rm nu}(x)}{\widehat{p}_{\rm de}(x)} = \left( \frac{5}{10^7} \right)^{-1} \left(\frac{1}{10^4} \right) = 200.
\end{align*}
Therefore, $r_{\rm MLE}(x)$ overestimates LRs even when the frequency $f_{\rm nu}(x)$ is lower than six.
At $f_{\rm nu}(x)=7$, $r_{\rm th}(x)$ is as high as 200, but suddenly drops to zero at $f_{\rm nu}(x)=6$.
In other words, $r_{\rm th}(x)$ only considers whether $f_{\rm nu}(x)$ is below the threshold to avoid overestimation.
Consequently, even when $f_{\rm nu}(x)=7$, which is slightly above the threshold, $r_{\rm th}(x)$ yields a high estimate.
By contrast, our estimator can achieve a conservative estimation based on low frequencies.
Consequently, from $f_{\rm nu}(x)=7$ to $10$, which is slightly above the threshold, the estimates of our estimator are much lower than $r_{\rm th}(x)=200$.

Our experiment used LR estimation to predict the word bigrams that occur to the left of named entities (NEs) from a corpus.
We show that our estimator can improve the prediction accuracy compared with $r_{\rm MLE}(x)$ and $r_{\rm th}(x)$.
Furthermore, to validate the effectiveness of the threshold, we measured the computation time and memory usage required to perform this task.
Consequently, we demonstrate that our estimator can perform efficient predictions.

\section{Related Work}
\label{sec:related}

A naive LR estimation approach estimates the probability distributions and takes their ratio.
As described in Section~\ref{sec:intro}, this approach yields unstable estimates and often overestimates LRs for low frequencies.
To overcome this problem, various ``direct LR estimation methods'' have been proposed to estimate LRs without estimating probability distributions~\cite{Huang:07,Bickel:07},~\cite{Sugiyama:08,Kanamori:09}.
However, these methods are used to estimate the LRs defined in continuous sample spaces and assume continuous values as sampled elements.
Therefore, Kikuchi et al.~\cite{Kikuchi:19} changed the basis functions used in unconstrained least-squares importance fitting (uLSIF)~\cite{Kanamori:09}, a direct estimation method based on least-squares fitting, to make uLSIF applicable to discrete sample spaces.
This estimator estimates LRs conservatively without a frequency threshold owing to the $\ell_2$-regularization introduced by optimization.
We also formulate our estimator using the basis functions of Kikuchi et al., but change the regularization scheme from $\ell_2$ to $\ell_1$.
This change introduced a frequency threshold.
Moreover, our estimator conservatively estimates LRs only for elements with frequencies higher than the threshold.
Kikuchi et al.~\cite{Kikuchi:19} aimed to provide conservative LR estimation for elements with all frequencies.
By contrast, our estimator aims to improve the efficiency of practical tasks that use LRs.

Statistical estimation approaches with frequency thresholds are simple and not only prevent overestimation due to low frequency but also improve the efficiency of practical tasks that use statistics.
Hence, these approaches were used to estimate various statistics~\cite{Agrawal:94, Montella:11}.
However, most of them only focus on setting thresholds and do not focus on the problem of overestimation owing to low frequencies near the thresholds.
Aoba et al.~\cite{Aoba:20} proposed a conservative estimator for conditional probability from low frequencies near a threshold.
This estimator is similar to ours in its derivation process and behavior.
Conditional probability is the ratio of probabilities, and can be interpreted as a type of LR.
Thus, our study result can be considered an extension of the results of Aoba et al. to general LR estimation.

\section{Our Estimator}
\label{sec:ours}

We then formulate our estimator.
This estimator conservatively estimates the LRs of low-frequency elements slightly above a frequency threshold and uniformly sets the LRs of the elements below the threshold to zero.

We describe the problem setting for LR estimation.
Let $D \subset \mathbb{U}$ be a set of discrete elements $x$s that a dataset contains, and $\mathbb{U}$ be a set of all $v$ types of elements that can exist, which is also known as a finite alphabet in information theory.
Suppose that we observe two independent and identically distributed (i.i.d.) samples:
\begin{align*}
\{x_{i}^{\rm de}\}_{i=1}^{n_{\rm de}} \overset{\rm i.i.d.}{\sim} p_{\rm de}(x), \quad \{x_{j}^{\rm nu}\}_{j=1}^{n_{\rm nu}} \overset{\rm i.i.d.}{\sim} p_{\rm nu}(x),
\end{align*}
where $x$ represents a language element such as a letter or word.
Following previous studies, we assume that:
\begin{align*}
p_{\rm de}(x) > 0 \quad {\rm for\ all\ } x \in D
\end{align*}
holds.
Under this assumption, we can define LRs for all $x$s.
This section discusses the estimation problem for the following LR:
\begin{align*}
r(x) = \frac{p_{\rm nu}(x)}{p_{\rm de}(x)}
\end{align*}
directly using the two samples $\{x_{i}^{\rm de}\}_{i=1}^{n_{\rm de}}$ and $\{x_{j}^{\rm nu}\}_{j=1}^{n_{\rm nu}}$ without performing a probability distribution estimation.
Indices ``de'' and ``nu'' represent the denominator and numerator of $r(x)$, respectively.

The uLSIF~\cite{Kanamori:09}, a direct estimation method based on least-squares fitting, defines the estimation model as follows:
\begin{align*}
\widehat{r}(x) = \sum_{l=1}^{b} \beta_l \varphi_l (x),
\end{align*}
where $\mbox{\boldmath $\beta$}=(\beta_1, \beta_2, \ldots ,\beta_{b})^{\mathrm{T}}$ are the parameters learned from the samples and $\{\varphi_l\}_{l=1}^{b}$ are basis functions that take non-negative values.
The original uLSIF estimates LRs defined in continuous spaces and uses Gaussian kernel-based basis functions.
However, we treat discrete elements, such as letters and words.
Their LRs were also defined in discrete spaces, and the Gaussian kernels were not effective in our study.
Therefore, we substitute the basis functions $\{\delta_l\}_{l=1}^{v}$ defined for each type of discrete element~\cite{Kikuchi:19}:
\begin{align}
\label{eq:phi}
\delta_l(x)= 
\begin{cases}
    1 & {\rm if\ }x = x_{(l)}, \\
    0 & {\rm otherwise},
      \end{cases}
\end{align}
where $l$ is an index that specifies the type of element, and $x_{(l)}$ is the $l$-th element of the $v$ types of elements that exist.
By using the basis functions in Eq. (\ref{eq:phi}), the estimation model for $x_{(m)},\ m = 1, 2, \ldots , v$ is as follows:
\begin{align}
\label{eq:linear_model}
\widehat{r}\left( x_{(m)} \right) = \sum_{l=1}^{v} \beta_l \delta_l \left( x_{(m)} \right) = \beta_m.
\end{align}
uLSIF finds the parameters $\mbox{\boldmath $\beta$}$ that minimize the squared error between the estimation model $\widehat{r}\left( x_{(m)} \right)$ and true LR $r\left( x_{(m)} \right)$.
To prevent overfitting, this method introduces $\ell_2$-regularization to squared-error minimization.
In contrast, we introduce $\ell_1$-regularization for efficient LR estimation.
Our optimization problem is as follows:
\begin{align}
\label{eq:uLSIF}
\min_{\mbox{\boldmath $\beta$} \in \mathbb{R}^v} \left[ \frac{1}{2} \mbox{\boldmath $\beta$}^{\mathrm{T}} \widehat{\mbox{\boldmath $H$}} \mbox{\boldmath $\beta$} - \widehat{\mbox{\boldmath $h$}}^{\mathrm{T}} \mbox{\boldmath $\beta$} + \lambda_{\rm L1} \sum_{l=1}^v \beta_l \right],
\end{align}
where $\mathbb{R}^v$ is a real $v$-dimensional space\footnote{See the original uLSIF paper~\cite{Kanamori:09} for the derivation of Eq. (\ref{eq:uLSIF}).
We changed the regularization scheme from $\ell_2$ to $\ell_1$.
}.
This equation introduces a penalty term $\lambda_{\rm L1} \sum_{l=1}^v \beta_l$ for the regularization of $\mbox{\boldmath $\beta$}$.
$\lambda_{\rm L1}\ (\ge 0)$ is the regularization parameter.
The $\ell_1$-regularization term in Eq.~(\ref{eq:uLSIF}) was originally $\sum_{l=1}^v |\beta_l|$.
As expressed in Eq. (\ref{eq:linear_model}), $\beta_l$ is the estimation model of $r(x)$, and because of the nonnegativity of $r(x)$, $\beta_l$ is also nonnegative.
Therefore, we can replace $|\beta_l|$ with $\beta_l$. 
$\widehat{\mbox{\boldmath $H$}}$ is a $v \times v$ matrix and its $(l,l')$-th element $\widehat{H}_{l,l'}$ is defined as follows:
\begin{align*}
\widehat{H}_{l,l'} &= \frac{1}{n_{\rm de}}\sum_{i=1}^{n_{\rm de}} \delta_l\left( x_i^{\rm de} \right) \delta_{l'}\left( x_i^{\rm de} \right)
= \begin{cases}
    \frac{f_{\rm de}\left( x_{(l)} \right)}{n_{\rm de}} & {\rm if\ }l=l', \\
    0 & {\rm otherwise}.
\end{cases}
\end{align*}
From the definition above, $\widehat{\mbox{\boldmath $H$}}$ is a diagonal matrix.
$\mbox{\boldmath $h$}$ is the $v$-dimensional vector and its $l$-th element $\widehat{h}_{l}$ is defined as follows:
\begin{align*}
\widehat{h}_l = \frac{1}{n_{\rm nu}} \sum_{j=1}^{n_{\rm nu}} \delta_l \left( x_j^{\rm nu} \right) = \frac{f_{\rm nu}\left( x_{(l)} \right)}{n_{\rm nu}},
\end{align*}
where $f_{*} \left( x_{(l)} \right)$ is the observed frequency of $x_{(l)}$ sampled from a probability distribution with density $p_{*}(x)$ $* \in \{{\rm de, nu}\}$.
In the objective function of Eq. (\ref{eq:uLSIF}), the first and second terms are derived from the squared error of $\widehat{r}(x)$ and true LR $r(x)$.
Using vector and matrix elements, we can represent the first and second terms as follows:
\begin{align}
\label{eq:uLSIF_1}
\frac{1}{2} \mbox{\boldmath $\beta$}^{\mathrm{T}} \widehat{\mbox{\boldmath $H$}} \mbox{\boldmath $\beta$} &= \frac{1}{2} \sum_{l=1}^v \frac{f_{\rm de}\left( x_{(l)} \right)}{n_{\rm de}} \beta_l^{2}, \\
\label{eq:uLSIF_2}
- \widehat{\mbox{\boldmath $h$}}^{\mathrm{T}} \mbox{\boldmath $\beta$} &= - \sum_{l=1}^v \frac{f_{\rm nu}\left( x_{(l)} \right)}{n_{\rm nu}} \beta_l.
\end{align}
We partially differentiated the objective function in Eq. (\ref{eq:uLSIF}) by $\beta_m$ and set it to zero:
\begin{align}
\label{eq:aaa_1}
\frac{\partial}{\partial \beta_m} \left( \frac{1}{2} \mbox{\boldmath $\beta$}^{\mathrm{T}} \widehat{\mbox{\boldmath $H$}} \mbox{\boldmath $\beta$} - \widehat{\mbox{\boldmath $h$}}^{\mathrm{T}} \mbox{\boldmath $\beta$} + \lambda_{\rm L1} \sum_{l=1}^v \beta_l \right) = 0.
\end{align}
Subsequently, by substituting Eqs. (\ref{eq:uLSIF_1}) and (\ref{eq:uLSIF_2}) into Eq. (\ref{eq:aaa_1}) and solving for $\beta_m$, we obtain the following parameter:
\begin{align}
\label{eq:aaa_2}
\widehat{r}\left( x_{(m)} \right) &= \widetilde{\beta}_m(\lambda_{\rm L1}) \nonumber \\
&= \left(\frac{f_{\rm de} \left( x_{(m)} \right)}{n_{\rm de}} \right)^{-1} \left(\frac{f_{\rm nu} \left( x_{(m)} \right)}{n_{\rm nu}} - \lambda_{\rm L1} \right)
\end{align}
that minimizes the squared error.
From Eq. (\ref{eq:linear_model}), $\widetilde{\beta}_m(\lambda_{\rm L1})$ is the estimation model $\widehat{r}\left( x_{(m)} \right)$.
Although $\widehat{r}\left( x_{(m)} \right)$ takes a negative value when $\lambda_{\rm L1}$ is greater than $\frac{f_{\rm nu} \left( x_{(m)} \right)}{n_{\rm nu}}$, $r(x)$ is always non-negative.
If the solution of Eq. (\ref{eq:uLSIF}) takes a negative value, the original uLSIF approximates it to zero.
Thus, when Eq. (\ref{eq:aaa_2}) takes a negative value, we round it to zero, according to the uLSIF framework.
Consequently, our estimator is defined as follows:
\begin{align*}
&r_{\rm L1}\left( x_{(m)} \right) \\
&= \left(\frac{f_{\rm de} \left( x_{(m)} \right)}{n_{\rm de}} \right)^{-1} \max \left(\frac{f_{\rm nu} \left( x_{(m)} \right)}{n_{\rm nu}} - \lambda_{\rm L1},\ 0 \right),
\end{align*}
setting the estimate to zero regardless of the other frequencies when the numerator $\frac{f_{\rm nu} \left( x_{(m)} \right)}{n_{\rm nu}}$ is less than or equal to $\lambda_{\rm L1}$.
On the contrary, if the numerator is higher than $\lambda_{\rm L1}$, this equation can yield a conservative estimate by subtracting $\lambda_{\rm L1}$ from the numerator.

The value of $\lambda_{\rm L1}$ can be set to any real number greater than or equal to zero.
We define $\lambda_{\rm L1}=\frac{\theta_{\rm L1}}{n_{\rm nu}}$ for a fair comparison between $r_{\rm L1}\left( x_{(m)} \right)$ and an estimator that simply sets a threshold.
This definition allowed us to set a frequency threshold.
Here, our estimator $r_{\rm L1}\left( x_{(m)} \right)$ is replaced by:
\begin{align}
\label{eq:ours}
r_{\rm L1}\left( x_{(m)} \right) =& \left(\frac{f_{\rm de} \left( x_{(m)} \right)}{n_{\rm de}} \right)^{-1} \frac{\max \left( f_{\rm nu} \left( x_{(m)} \right) - \theta_{\rm L1},\ 0 \right)}{n_{\rm nu}},
\end{align}
where $\theta_{\rm L1}\ (\ge 0)$ denotes the frequency threshold for $f_{\rm nu} \left( x_{(m)} \right)$.
The estimate becomes highly conservative when $f_{\rm nu} \left( x_{(m)} \right)$ is slightly higher than $\theta_{\rm L1}$.
In addition, this estimator eliminates LR estimation for frequencies below the threshold, making practical tasks using LRs more efficient.
When $\theta_{\rm L1}$ is zero, the estimator is equal to the ratio of the relative frequencies of the probability distributions of the denominator and numerator of $r\left( x_{(m)} \right)$.

\section{Experiment}
\label{sec:experiments}

\begin{figure}[tb]
  \centering
  \includegraphics[keepaspectratio, scale=0.46]{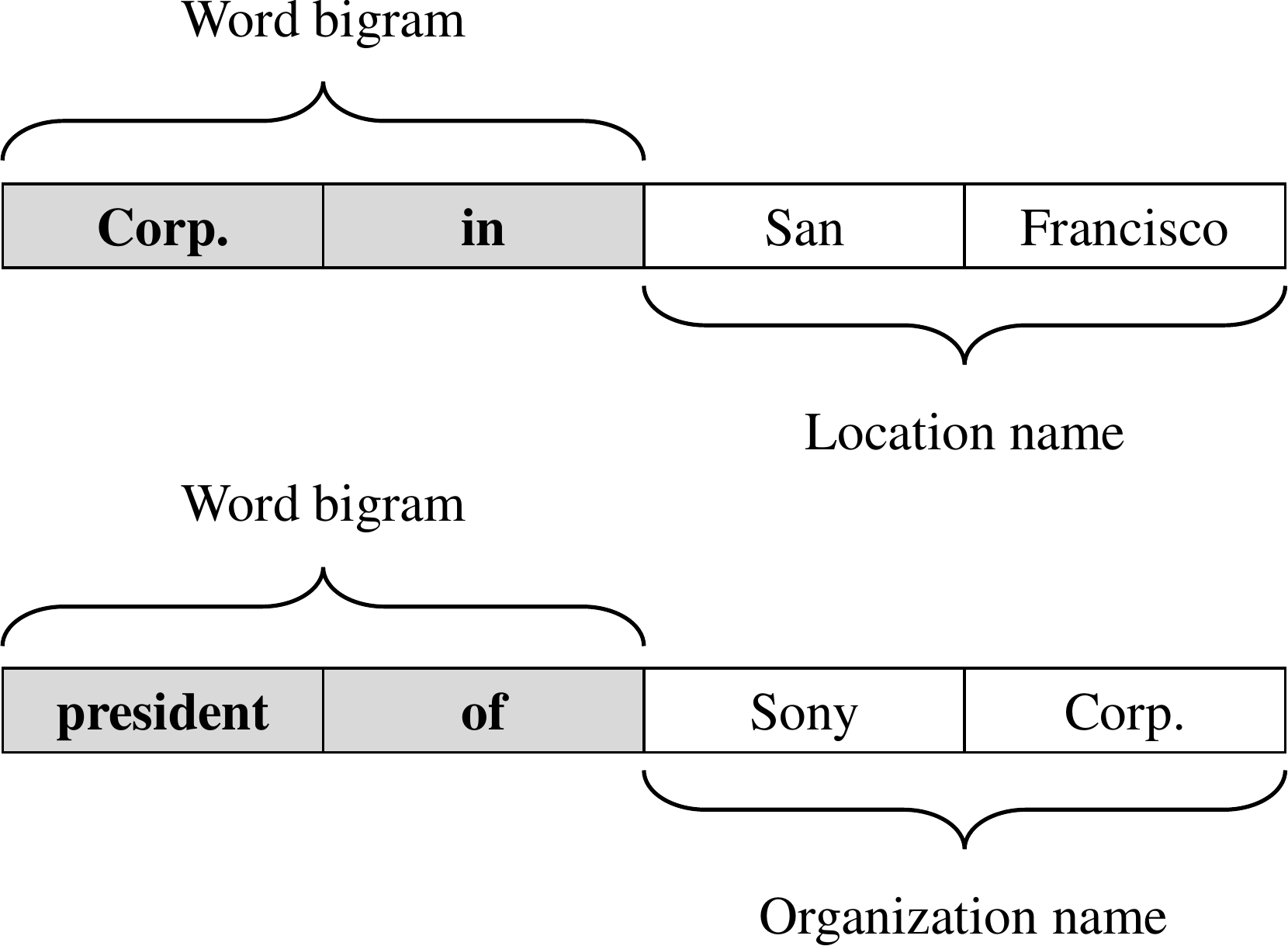}
  \caption{Contexts for two types of NEs}
  \label{fig:contexts}
\end{figure}
We predict from a corpus the contexts in which NEs occur on the left using LRs.
In our experiment, we evaluated the effectiveness of our estimator in terms of context prediction accuracy and prediction efficiency.
As shown in Fig. \ref{fig:contexts}, we use location and organization names (tagged as LOC and ORG, respectively) as the NEs and the word bigrams to the left of those names as the contexts.
This context can be predicted for two reasons.
First, bigrams are abundant in variety, but most are infrequent.
Here, the prediction accuracy and efficiency vary significantly depending on how the infrequent bigrams are handled.
Therefore, we can clarify the differences among LR estimators that handle low frequencies differently, and validate the effectiveness of our estimator.
Second, the occurrence contexts were uniquely determined, allowing quantitative evaluations of the estimators.

\subsection{Experimental Environment}

The experimental environment is described as follows:
\begin{itemize}
    \item OS : Windows 10 Home
    \item processor : AMD Ryzen 7 Extreme Edition @1.80GHz
    \item memory : 16.0GB
    \item Python : 3.6.10
\end{itemize}

\subsection{Experimental Datasets and Conditions}

\begin{table}[tb]
\caption{Information of experimental datasets.
Each column represents the number of types or total frequency of bigrams.}
\label{tab:datasets}
\centering
	\begin{tabular}{ c  r  r  r  r  } \hline
		 \multicolumn{1}{ c }{\multirow{2}{*}{Data}} & \multicolumn{2}{ c }{All articles} & \multicolumn{2}{ c }{LOC} \\ \cline{2-5}
		 & \multicolumn{1}{ c }{Types} & \multicolumn{1}{ c }{Freq.} & \multicolumn{1}{ c }{Types} & \multicolumn{1}{ c }{Freq.} \\ \hline
		Train & 1,468,292 & 4,002,930 & 31,294 & 64,072 \\
		Valid & 230,528 & 401,445 & 4,318 & 6,116 \\
		Eval & 231,931 & 403,145 & 4,164 & 5,876 \\ \hline
		\multicolumn{1}{ c }{\multirow{2}{*}{Data}} & \multicolumn{2}{ c }{ORG} \\ \cline{2-3}
		& \multicolumn{1}{ c }{Types} & \multicolumn{1}{ c }{Freq.} \\ \cline{1-3}
		Train & 44,946 & 94,737 \\
		Valid & 6,443 & 9,946 \\ 
		Eval & 6,544 & 9,857 \\ \cline{1-3}
	\end{tabular}
\end{table}
We used the following procedure to create experimental datasets based on the 1987 edition of the Wall Street Journal Corpus \footnote{\url{https://catalog.ldc.upenn.edu/LDC2000T43}}.
First, we randomly sampled 12,000 articles from the corpus.
We then used the Stanford named entity recognizer (Stanford NER)~\cite{Finkel:05}\footnote{\url{https://nlp.stanford.edu/software/CRF-NER.html}} to assign NE tags to the sampled articles.
We allocated 10,000 articles for training, 1,000 for validation, and 1,000 for evaluation.
We divided the training articles into word bigrams and counted the frequencies required for LR estimation.
Frequency information was used as the training dataset.
We used the set of all bigrams in the validation and evaluation articles as validation and evaluation datasets, respectively.
We used the NE tags of the validation and evaluation articles only for correct and incorrect judgment.
Table~\ref{tab:datasets} lists the experimental datasets.
There is only a 1.5- to 2-fold difference in the number of types and total frequency for bigrams in each dataset, indicating several low-frequency bigrams.

We specified the types of NEs as the experimental conditions.
We have two choices: LOC and ORG, and we select one of them.

\subsection{Experimental Procedure}

First, we store the frequencies of the bigrams in memory from the training dataset.
We do not store frequencies below the threshold for estimators with a threshold because they are unnecessary for LR estimation.
Next, we estimate
\begin{align}
\label{eq:estimation_target}
r(x) = \frac{p(x | c_{\rm NE})}{p(x)}
\end{align}
for all bigrams $x$s in the evaluation dataset.
$c_{\rm NE}$ denotes the class label assigned to the left bigrams of NEs.
$p(x | c_{\rm NE})$ represents the occurrence probability of $x$ to the left of NEs and $p(x)$ is the occurrence probability of $x$ in the training articles.

For each estimator, we sort bigrams in descending order of the estimates and classify the top 4,000 bigrams as correct or incorrect.
If a bigram occurred once to the left of NEs in the evaluation articles, we classified it as correct (context); otherwise, we classified it as incorrect.
Using the classification results, we plotted rank--recall curves for each estimator described in Section \ref{sec:comparison_method}.
The curves were plotted on a graph with the rank of a bigram on the horizontal axis and the recall at the rank on the vertical axis.
The estimator with the highest recall at a rank is the best at that rank.
In this graph, the slope of the line connecting the point of a rank on the curve and the origin of the graph is proportional to the precision of the rank.
The recall and precision are defined as follows:
\begin{align*}
\frac{|\{x \mid x \in R \cap X\}|}{|R|}\ {\rm and}\ 
\frac{|\{x \mid x \in R \cap X\}|}{|X|},
\end{align*}
respectively, where $X$ denotes the set of the top 4,000 bigrams and $R$ is the set of bigrams that occur to the left of NEs in the evaluation articles, that is, the set of right bigrams.

We also measured the computation time and memory usage to confirm the improvement in efficiency by setting a threshold.
We define the computation time as the time required to store the frequency of the bigrams used for training and to estimate the LR of all the bigrams in the evaluation dataset.
We used the average of 10 repetitions of the experimental procedure as the computation time.
Memory usage is the amount of memory required to store all frequencies of the bigrams used for training.

\subsection{Comparison Estimators}
\label{sec:comparison_method}

We compared the following four LR estimators.
Each estimator estimates $r(x)=\frac{p_{\rm nu}(x)}{p_{\rm de}(x)}$.
As expressed in Eq. (\ref{eq:estimation_target}), $p_{\rm de}(x)$ and $p_{\rm nu}(x)$ correspond to $p(x)$ and $p(x | c_{\rm NE})$, respectively.
Estimators 1 and 2 do not have a threshold, but estimators 3 and 4 do.

\noindent
\textbf{1: Baseline}\quad
$r_{\rm MLE}(x)$ estimates the two probability distributions as relative frequencies and uses their ratio.
This estimator is defined by Eq.~(\ref{eq:MLE}) in Section~\ref{sec:intro}.
If the frequency of a bigram is zero in the training dataset, we cannot compute $r_{\rm MLE}(x)$ because of the zero division.
In this case, we regard the estimate as zero.

\noindent
\textbf{2: L2}\quad
$\ell_2$-regularization provides a conservative estimate depending on the low frequencies.
$r_{\rm L2}(x)$ is defined as follows:
\begin{align*}
& r_{\rm L2}(x) = \left(\frac{f_{\rm de} (x)}{n_{\rm de}} + \lambda_{\rm L2} \right)^{-1} \frac{f_{\rm nu}(x)}{n_{\rm nu}},
\end{align*}
where $\lambda_{\rm L2}\ (\ge 0)$ is the regularization parameter.
This estimator is derived from the minimization framework of squared error with $\ell_2$-regularization.
Because $r_{\rm L2}(x)$ does not have a frequency threshold, the efficiency of solving the experimental task is lower than that of our estimator.

\noindent
\textbf{3: Threshold}\quad
$r_{\rm th}(x)$ is a simple threshold approach defined in Eq.~(\ref{eq:th}) in Section~\ref{sec:intro}.

\noindent
\textbf{4: L1 (Ours)}\quad
This is the proposed estimator and is defined in Eq.~(\ref{eq:ours}) in Section~\ref{sec:ours}.

Estimators 2--4 have hyperparameters $\lambda_{\rm L2}$, $\theta_{\rm th}$, and $\theta_{\rm L1}$, respectively.
We determined the parameter values of ``L2'' and ``Threshold'' using the following procedure:
First, we considered the validation dataset as the evaluation dataset and plotted rank--recall curves for each parameter value of each estimator.
We then set the value with the largest area under the curve as the optimal value.
In ``L2,'' we varied $\lambda_{\rm L2}$ to $10^{-9}, 10^{-8}, \ldots ,10^{-1}$ and set $10^{-2}$ as the optimal values for both LOC and ORG.
In the ``Threshold,'' we varied $\theta_{\rm th}$ to $9, 8, \ldots, 1$ and set $2$ as the optimal values for both LOC and ORG.
To make an equal comparison with ``Threshold,'' in ``L1'' (our estimator), we set $\theta_{\rm L1}$ to 2 as well as $\theta_{\rm th}$.
We also investigated our estimator's parameter value that maximizes the area under the rank--recall curve.
Consequently, we found that the optimal value of $\theta_{\rm L1}$ was 2 as well as $\theta_{\rm th}$.

\subsection{Experimental Results}

\begin{table*}[tb]
\caption{Computation time and memory usage. Computation time is the average of 10 times.}
\label{tab:Time-Memory}
	\begin{minipage}[b]{0.495 \linewidth}
		\centering
		\subcaption{LOC}\label{tab:Time-Memory_LOCATION}
		\begin{tabular}{ l r r r } \hline
			 \multicolumn{1}{ c }{Estimator} & \multicolumn{1}{ c }{Threshold} & \multicolumn{1}{ c }{Time [sec]}  &  \multicolumn{1}{ c }{Memory [KB]} \\ \hline
			 Baseline & None & 3.179 & 8,180 \\
			 L2 & None & 3.179 & 8,180 \\
			 Threshold & 2 & 0.822 & 854 \\
			 L1 (Ours) & 2 & 0.823 & 854 \\ \hline
		\end{tabular}
	\end{minipage}
	\begin{minipage}[b]{0.495 \linewidth}
		\centering
		\subcaption{ORG}\label{fig:Time-Memory_ORGANIZATION}
		\begin{tabular}{ l r r r } \hline
			 \multicolumn{1}{ c }{Estimator} & \multicolumn{1}{ c }{Threshold} & \multicolumn{1}{ c }{Time [sec]}  &  \multicolumn{1}{ c }{Memory [KB]} \\ \hline
			 Baseline & None & 3.518 & 13,292 \\
			 L2 & None & 3.289 & 13,292 \\
			 Threshold & 2 & 0.838 & 1,015 \\
			 L1 (Ours) & 2 & 0.836 & 1,015 \\ \hline
		\end{tabular}
	\end{minipage}
\end{table*}

\begin{figure*}[tb]
	\begin{minipage}[b]{0.495 \linewidth}
		\centering
		\includegraphics[keepaspectratio, scale=0.55]{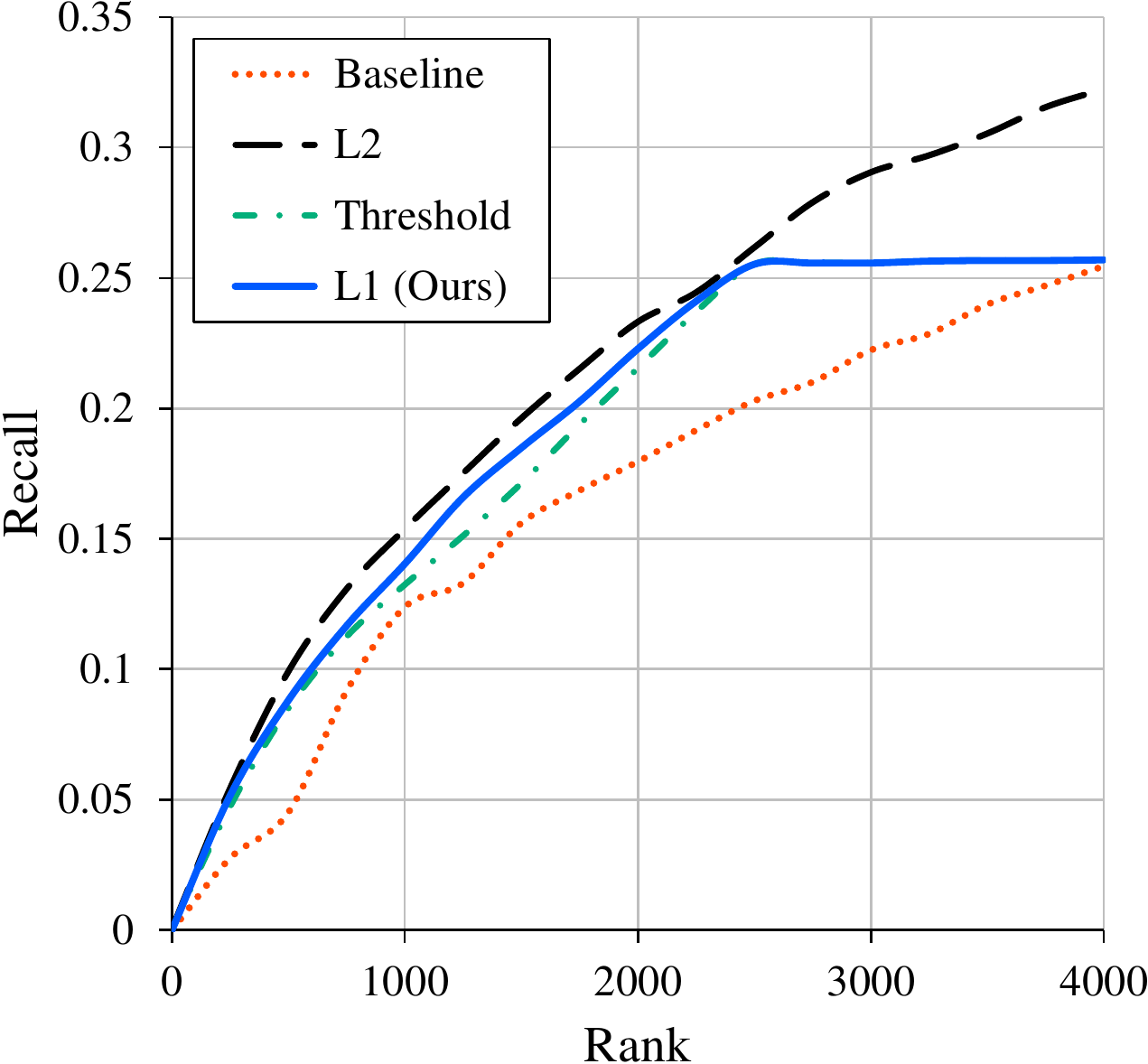}
		\subcaption{LOC}\label{fig:Rank_Recall_LOCATION}
	\end{minipage}
	\begin{minipage}[b]{0.495 \linewidth}
		\centering
		\includegraphics[keepaspectratio, scale=0.55]{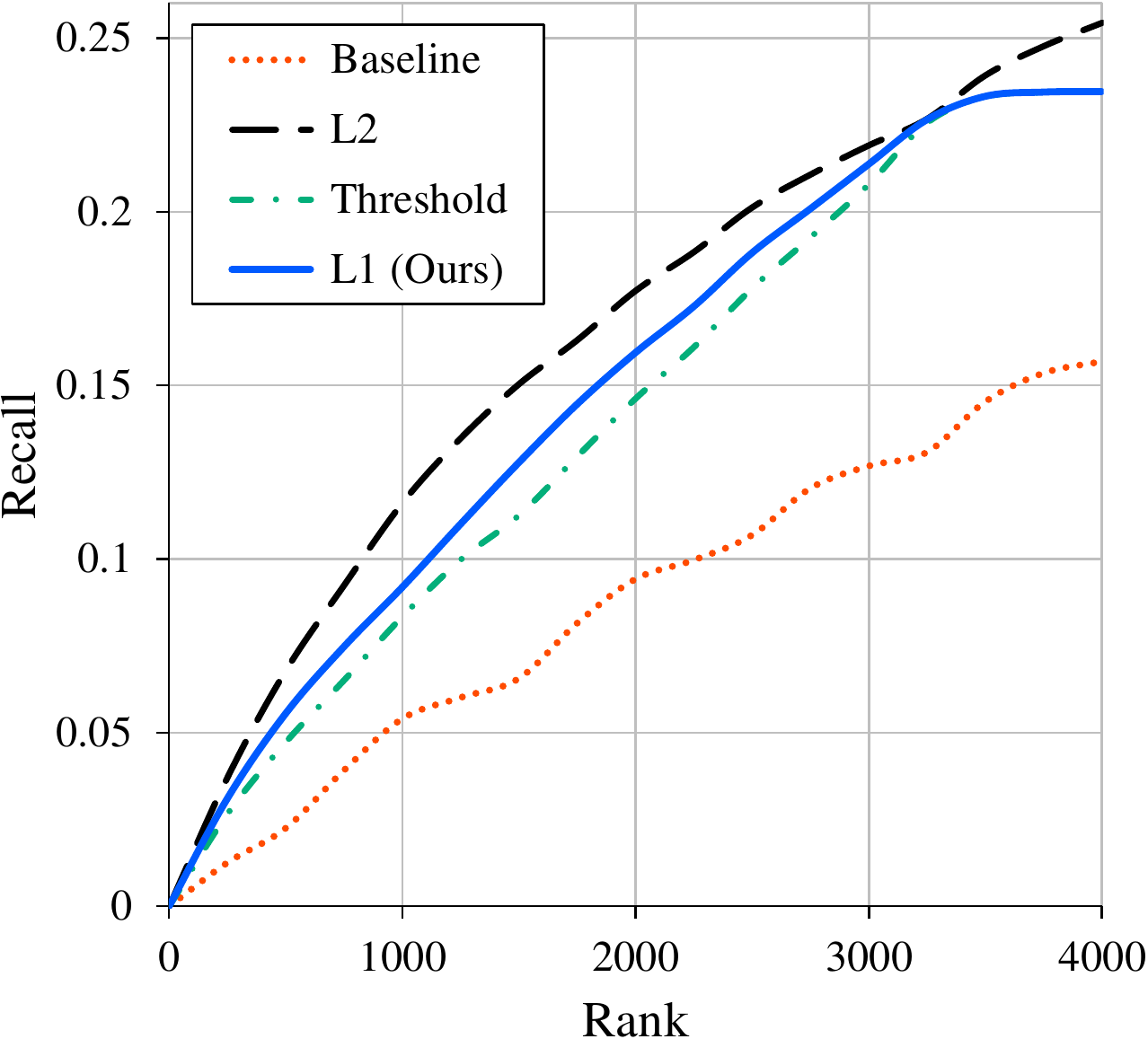}
		\subcaption{ORG}\label{fig:Rank_Recall_ORGANIZATION}
	\end{minipage}
	\caption{Rank--recall curves}\label{fig:Rank--Recall}
\end{figure*}
Fig.~\ref{fig:Rank--Recall} shows the rank--recall curves.
The curves were plotted on a graph with the rank of a bigram on the horizontal axis and the recall at the rank on the vertical axis.
The estimator with the highest recall at a rank is the best at that rank.
The slope of the line connecting the point of a rank on the curve and the origin of the graph is proportional to the precision of the rank.
``Baseline'' has poor prediction accuracy compared to other estimators because this estimator overestimates the LRs of low-frequency bigrams and depreciates the reliable bigrams that occur frequently.
In particular, the left bigrams of ORG are rich in variety but infrequent, which emphasizes the poor performance of ``Baseline,'' as shown in Fig. \ref{fig:Rank--Recall} (b).
``L2'' conservatively estimates the LRs of low-frequency bigrams and rank frequent bigrams to be higher.
Therefore, this estimator achieved the best performance for both LOC and ORG.
Estimators with thresholds, ``Threshold'' and ``L1 (Ours),'' outperform ``Baseline'' because the thresholds prevent overestimating the LRs.
In addition, we observe that ``L1'' is slightly better than ``Threshold.''
Thus, the results suggest that conservative LR estimation near the threshold is more effective than a simple threshold approach.
However, these estimators uniformly set all LRs for bigrams below the thresholds to zero.
This effect results in accuracy degradation compared to ``L2,'' with the negative influence that recall does not improve at the lower rank.

Table~\ref{tab:Time-Memory} lists the computation time and memory usage.
First, we compare the estimators with and without the thresholds.
These comparisons show that setting a threshold improves the efficiency by approximately 1/4 in computation time and 1/10 in memory usage.
Next, we compare the estimators with thresholds, ``Threshold'' and ``L1.''
Although ``L1'' has an additional process that subtracts the frequency by the threshold for conservative estimation, there is almost no difference in computation time and memory usage between the two estimators.
These results suggest the effectiveness of ``L1'' in terms of efficiency in solving context prediction tasks.

\section{Conclusion}

We propose a conservative LR estimator for low frequencies near the threshold frequency.
This estimator provides a frequency threshold for efficient estimation and suppresses the overestimation of LRs for low frequencies slightly above the threshold.
We derive an estimator using a theoretical framework that minimizes the squared error.
This framework provides conservative estimation using $\ell_1$-regularization.
We experimented by predicting the occurrence contexts of NEs from a corpus using LR and verified the effectiveness of our estimator in terms of prediction accuracy and efficiency.
The results suggest that our estimator improves the prediction accuracy compared to the simple threshold approach.
Moreover, they suggested that a threshold improves the prediction efficiency by approximately 1/4 of the computation time and 1/10 of the memory usage.
However, the performance difference between our estimator and the simple threshold estimation method was minimal.
Our future work is to verify the effectiveness of our estimator in tasks where a threshold is essential and the difference between the two estimators is clear.

\section*{Acknowledgment}

This work was supported in part by JSPS KAKENHI Grant Numbers JP19K12266, JP22K18006.

\bibliography{references}
\bibliographystyle{unsrt}

\end{document}